\documentclass[a4paper]{article}
\usepackage{idsiatechrep}      % english acronyms
\usepackage{graphicx}
\usepackage{natbib}
\usepackage{bbm}

\title{Handwritten Digit Recognition with a Committee of Deep Neural Nets on GPUs}
\author{Dan C. Cire\c{s}an, Ueli Meier, Luca M. Gambardella and J{\"u}rgen Schmidhuber}
\date{March 2011}

\reportnumber{03-11}

\reportaddnote{This work was partially supported by the Swiss Commission for Technology and Innovation (CTI), Project n. 9688.1 IFF: Intelligent Fill in Form.}

\begin{document}
\makecover         % makes the cover sheet
\maketitle

\begin{abstract}
The competitive MNIST handwritten digit recognition benchmark has a long history of broken records since 1998. The most recent substantial improvement by others dates back 7 years (error rate 0.4\%) . Recently we were able to significantly improve this result, using graphics cards to greatly speed up training of simple but deep MLPs, which achieved 0.35\%, outperforming all the previous more complex methods. Here we report another substantial improvement: 0.31\% obtained using a committee of MLPs. 
\end{abstract}

%\begin{keywords}
%  neural networks, multilayer perceptron, graphics processing unit, training set deformations 
%\end{keywords} 

\section{Introduction}
Current automatic handwriting recognition algorithms are already pretty good at learning to recognize handwritten digits.  More than a decade ago, Multilayer Perceptrons or MLPs \citep{Werbos:74, LeCun:85, Rumelhart:86} were among the first classifiers tested on the now famous MNIST handwritten digit recognition benchmark. Most had few layers or few artificial neurons (units) per layer \citep{LeCun:98}, but apparently back then these were the biggest feasible MLPs, trained when CPU cores were at least 20 times slower than today. A more recent MLP with a single hidden layer of 800 units achieved 0.70\% error \citep{Simard:03}. The latest substantial improvement by others occurred in 2003 \citep{Simard:03} (error rate 0.4\%). It was obtained with a convolutional neural network (CNN), using then novel elastic training image deformations. \cite{Ranzato:06, Ranzato:07} pre-trained each hidden CNN layer one by one in unsupervised fashion (this seems promising especially for small training sets), then used supervised learning to achieve 0.39\% error rate. Recently we were able to significantly improve this result, obtaining an error rate of 0.35\% using graphics cards (GPUs) to greatly speed up training of plain but deep MLPs \citep{Ciresan:2010}. Deformations proved essential to prevent MLPs with up to 12 million free parameters from overfitting. To let the deformation process  keep up with network training speed we had to  port it  onto the GPU as well. 

At some stage in the classifier design process one usually has collected a set of possible classifiers. Often one of them yields best performance. Intriguingly, however, the sets of patterns misclassified by the different classifiers do not necessarily overlap. This information could be harnessed in a committee. In the context of handwritten recognition, \cite{Chellapilla:06} already showed how a combination of various classifiers can be trained more quickly than a single classifier yielding the same error rate. Here we focus on improving recognition rate using a committee of MLPs. Our goal is to produce a group of classifiers whose errors on various parts of the training set differ as much as possible \citep{Bishop:06}. We show that for handwritten digit recognition this can be achieved by training identical classifiers on data normalized in different ways prior to training.

\section{Distorting images to get more training instances}
\label{sec:distortions}
MNIST consists of two datasets, one for training (60,000 images) and one for testing (10,000 images). Many studies divide the training set into two sets consisting of 50,000 images for training and 10,000 for validation. So far, however, the best results on MNIST were obtained by deforming training images, thus greatly increasing their number. This allows for training networks with many weights, making them insensitive to in-class variability. Our network is also trained on numerous slightly deformed images, continually generated in online fashion; hence we may use the whole un-deformed training set for validation, without wasting training images. Pixel intensities of the original gray scale images range from 0 (background) to 255 (max foreground intensity). $28\times28=784$ pixels per image get mapped to real values $\frac{pixel~intensity}{127.5}-1.0$  in $[-1.0,1.0]$, and are fed into the NN input layer.

 We combine affine (rotation, scaling and horizontal shearing) and elastic deformations, characterized by the following real-valued parameters:
\begin{itemize}
\item $\sigma$ and $\alpha$: for elastic distortions emulating uncontrolled oscillations of hand muscles \citep{Simard:03}. They are obtained by applying a displacement field to each digit. The displacement field is built by convolving a randomly initialized field with a Gaussian kernel whose standard deviation is defined by $\sigma$. $\alpha$ is a scaling factor controlling the amplitude of the applied elastic deformations;
\item $\beta$: a random angle from $[-\beta,+\beta]$ describes either rotation or horizontal shearing. In case of shearing, $\tan\beta$ defines the ratio between horizontal displacement and image height;
\item $\gamma_x$, $\gamma_y$: for horizontal and vertical scaling, randomly selected from $[1-\gamma/100,1+\gamma/100]$.
\end{itemize}

At the beginning of every epoch the entire MNIST training set gets deformed. Initial experiments with small networks suggested the following deformation parameters: $\sigma=5.0-6.0$, $\alpha=36.0-38.0$, $\gamma=15-20$. Since digits 1 and 7 are similar they get rotated/sheared less ($\beta=7.5^\circ$) than other digits ($\beta=15.0^\circ$).

\begin{figure}[ht]
\hfill
\begin{center}
\includegraphics[width=0.8\textwidth]{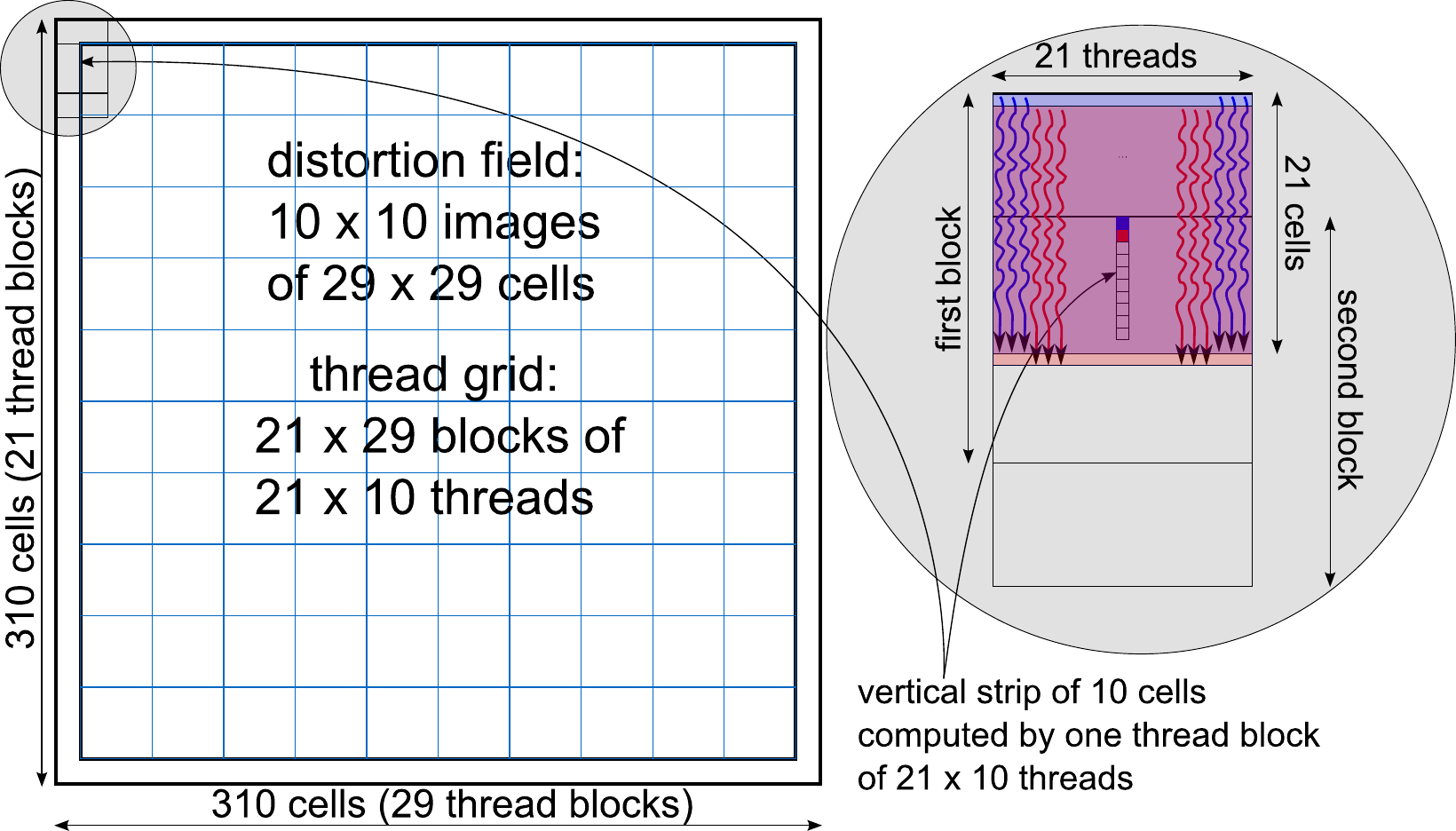}
\end{center}
\caption{Mapping the thread grid of convolution onto the distortion field.}
\label{Fig:Conv}
\end{figure}

It takes 83 CPU seconds to deform the 60,000 MNIST training images, most of them (75 seconds) for elastic distortions. Only the most time-consuming part of the latter---convolution with a Gaussian kernel---is ported to the GPU. The MNIST training set is split into 600 sequentially processed batches. MNIST digits are scaled from the original $28\times28$ pixels to $29\times29$ pixels, to get a proper center, which simplifies convolution. Each batch grid ($10\times10$ images) has 290 $\times$ 290 cells, zero-padded to 310 $\times$ 310, thus avoiding margin effects when applying a Gaussian convolution kernel of size $21\times21$.

The GPU program groups many threads into a block, where they share the same Gaussian kernel and parts of the random field. All 29 $\times$ 290 blocks contain 21 (the kernel size) $\times$10 threads, each computing a vertical strip of the convolution (Figure~\ref{Fig:Conv}).

Generating the elastic displacement field takes only 1.5 seconds. Deforming the whole training set is more than 10 times faster, taking 6.5 instead of the original 83 seconds. Further optimizations would be possible by porting all deformations onto GPU, and by using the hardware's interpolation capabilities to perform the final bilinear interpolation. We omitted these since deformations are already pretty fast (deforming all images of one epoch takes only 3-10 \% of total computation time, depending on MLP size).

\section{Forming a committee}

Preprocessing of the original MNIST data is mainly motivated by practical experience. MNIST digits are normalized such that the width or height of the bounding box equals 20 pixels. The variation of the aspect ratio for various digits is quite large, and we normalize the width of the bounding box to range from 10 to 20 pixels with a step-size of 2 pixels prior to training for all digits except ones. Normalizing the original MNIST training data results in 6 normalized training sets. In total we perform experiments with seven different data sets (6 normalized and the original MNIST).

\begin{figure}[ht]
\hfill
\begin{center}
\includegraphics[width=0.8\textwidth]{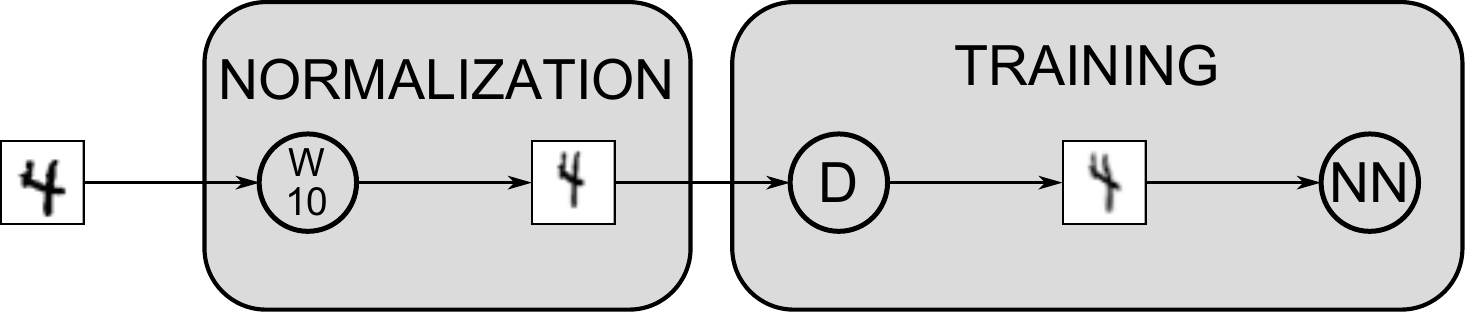}
\end{center}
\caption{Training a committee member. Original MNIST training data (left digit) is normalized (W10) prior to training (middle digit). The normalized data is distorted (D) for each training epoch and used as input (right digit) to the network (NN). Each depicted digit represents the whole training set.}
\label{Fig:training}
\end{figure}

The training procedure of a network is summarized in Figure \ref{Fig:training}. Each network is trained separately on normalized or original data. The normalization is done for all digits in the training set prior to training (normalization stage). During each training epoch the digits are distorted (Sec.~\ref{sec:distortions}) in a different way. For the network trained on original MNIST data the normalization step is omitted. 

\begin{figure}[ht]
\hfill
\begin{center}
\includegraphics[width=0.8\textwidth]{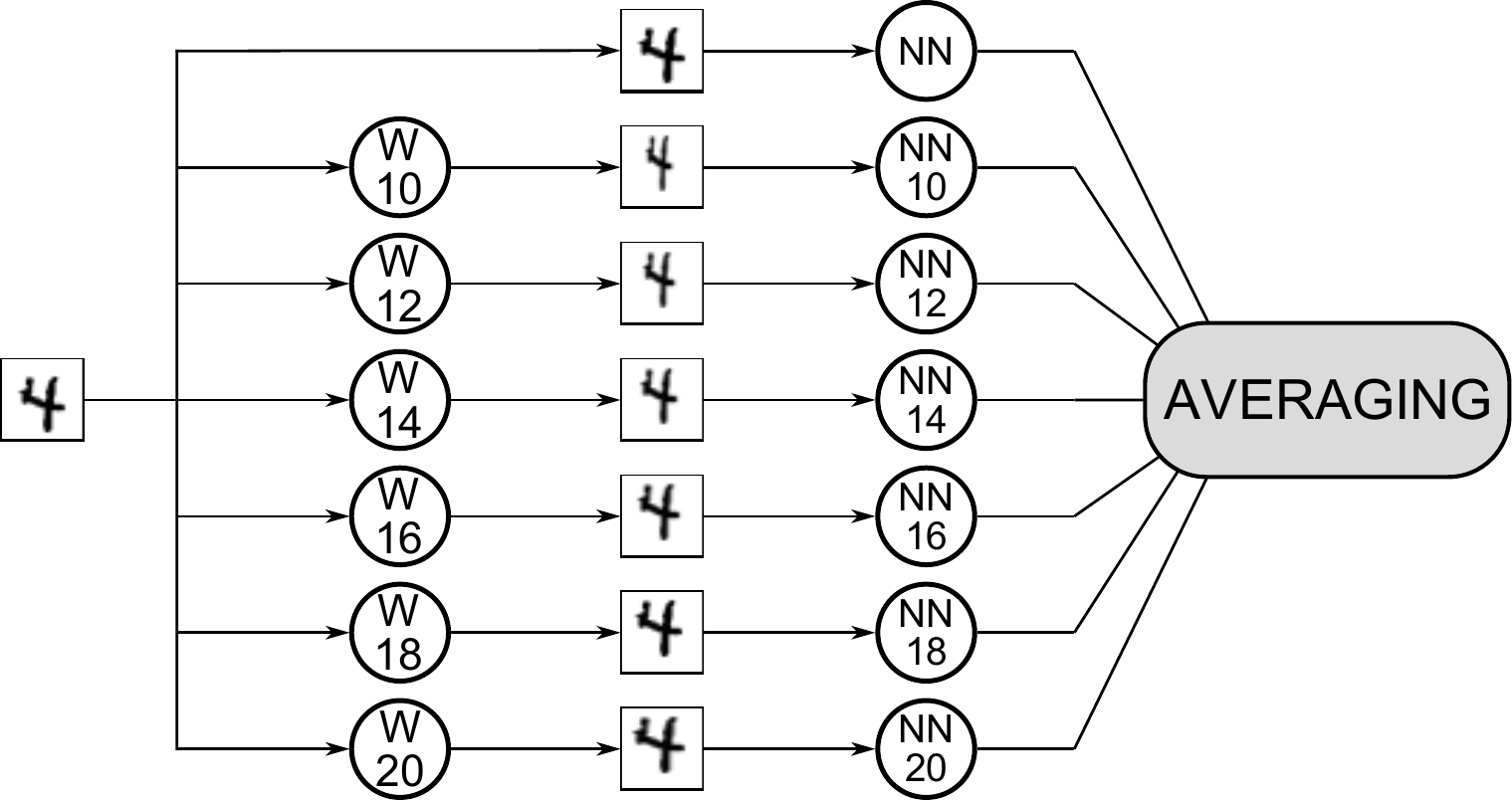}
\end{center}
\caption{Testing with a committee. If required, the input digits are width-normalized (W blocks) and then processed by the corresponding MLP. The committee is formed by averaging the outputs of all MLPs.}
\label{Fig:testing}
\end{figure}

We perform six experiments to analyze performance improvements due to committees. Each committee consists of seven randomly initialized one-hidden-layer MLPs with 800 hidden units, trained with the same algorithm on randomly selected batches. The five committees differ only in how the data are preprocessed (or not) prior to training and on how the data are deformed during training. The committees are formed by simply averaging the corresponding outputs as shown in Figure \ref{Fig:testing}. 

The first two experiments are performed on undeformed original MNIST images. We train a committee of seven MLPs on original MNIST and we also form a committee of MLPs trained on preprocessed data. In Table \ref{table:results_orig} the error rates are listed for each of the individual nets and the committees. The improvement of the committee with respect to the individual nets is marginal for the first experiment. Adding preprocessing, the individual experts as well as the corresponding committee of the second experiment achieve substantially better recognition rates.
 
\begin{table}
\caption{\textit{Error rates of individual nets and of the two resulting committees. For experiment 1 seven nets are trained on the original MNIST, whereas for experiment 2 seven nets are trained on preprocessed data: WN x - Width Normalization of the bounding box to be x pixels wide; ORIG - original MNIST.}}
\label{table:results_orig}
\begin{center}
\begin{tabular}{l || l r | l r}
&\multicolumn{4}{c}{Error rate [\%]} \\
&\multicolumn{2}{c}{Exp. 1}
&\multicolumn{2}{c}{Exp. 2}\\
\hline
\hline
Net 1: & init 1: & 1.79 & WN 10: & 1.62\\
Net 2: & init 2: & 1.80 & WN 12: & 1.37\\
Net 3: & init 3: & 1.77 & WN 14: & 1.48\\
Net 4: & init 4: & 1.72 & WN 16: & 1.53\\
Net 5: & init 5: & 1.91 & WN 18: & 1.56\\
Net 6: & init 6: & 1.86 & WN 20: & 1.49\\
Net 7: & init 7: & 1.75 & ORIG: & 1.79\\
\hline
Average:
 &\multicolumn{2}{c}{\bf{1.70}}
 &\multicolumn{2}{c}{\bf{1.31}}\\
\end{tabular}
\end{center}
\end{table}

To study the combined effect of preprocessing and deformation, we perform four additional experiments on deformed MNIST (Tab. \ref{table:results}). Unless stated otherwise, default elastic deformation parameters $\sigma=6$ and $\alpha=36$ are used. All experiments with deformed images use independent horizontal and vertical scaling of maximum $12.5\%$ and a maximum rotation of $\pm12.5^\circ$. 

Experiment 3 is similar to Experiment 1, except that the data are continually deformed. Error rates of the individual experts are much lower than without deformation (Tab. \ref{table:results_orig}).

In experiment 4 we randomly reselect training and validation sets for each of the individual experts, simulating in this way the bootstrap aggregation technique \citep{Breiman:96}. The resulting committee performs slightly better than that of experiment 3. 

In experiment 5 we vary deformations for each individual network. Error rates of some of the nets are bigger than in experiments 3 and 4, but the resulting committee has a lower error rate.

In the last experiment we train seven MLPs on preprocessed images that are also continually deformed during training. The error rate of the committee ($0.43$ \%) is the best result ever reported for such a simple architecture. 

\begin{table}
\caption{\textit{Error rates of the individual nets and of the resulting committees. In experiments 3 and 4 seven nets are trained on deformed ($\sigma=6$, $\alpha=36$) MNIST, whereas in experiment 4 training and validation sets are reselected. In experiment 5 seven nets are trained on deformed (different $\sigma$, $\alpha$) MNIST, and in experiment 6 seven nets are trained on normalized, deformed ($\sigma=6$, $\alpha=36$) MNIST. WN x - Width Normalization of the bounding box to be x pixels wide; ORIG - original MNIST.}}
\label{table:results}
\begin{center}
\begin{tabular}{l || l | c | c || l r | l r}
&\multicolumn{7}{c}{Error rate [\%]} \\
&\multicolumn{1}{c}{}
&\multicolumn{1}{c}{Exp. 3}
&\multicolumn{1}{c}{Exp. 4}
&\multicolumn{2}{c}{Exp. 5}
&\multicolumn{2}{c}{Exp. 6}\\
\hline
\hline
Net 1: & init 1: & 0.72 & 0.68 &$\sigma=4.5$ $\alpha=36$: & 0.69 & WN 10: & 0.64\\
Net 2: & init 2: & 0.71 & 0.82 &$\sigma=4.5$ $\alpha=42$: & 0.94 & WN 12: & 0.78\\
Net 3: & init 3: & 0.72 & 0.73 &$\sigma=6.0$ $\alpha=30$: & 0.55 & WN 14: & 0.70\\
Net 4: & init 4: & 0.71 & 0.69 &$\sigma=6.0$ $\alpha=36$: & 0.72 & WN 16: & 0.60\\
Net 5: & init 5: & 0.62 & 0.71 &$\sigma=6.0$ $\alpha=42$: & 0.60 & WN 18: & 0.59\\
Net 6: & init 6: & 0.65 & 0.70 &$\sigma=7.5$ $\alpha=30$: & 0.86 & WN 20: & 0.70\\
Net 7: & init 7: & 0.69 & 0.75 &$\sigma=7.5$ $\alpha=36$: & 0.79 & ORIG:  & 0.71\\
\hline
Average:
 &\multicolumn{1}{c}{}
 &\multicolumn{1}{c|}{\bf{0.56}}
 &\multicolumn{1}{c||}{\bf{0.53}}
 &\multicolumn{2}{r|}{\bf{0.49}}
 &\multicolumn{2}{r}{\bf{0.43}}\\
\end{tabular}
\end{center}
\end{table}

\section{Experiments on GPUs}

All simulations were performed on a computer with a Core i7 920 2.66GHz processor, 12GB of RAM, and a GTX 480 graphics card. The GPU accelerates the deformation routine by a factor of 10 (only elastic deformations are GPU-optimized); the forward propagation (FP) and BP routines are sped up by a factor of 50. We pick the trained MLP with the lowest validation error, and evaluate it on the MNIST test set.

\begin{figure}[ht]
\hfill
\begin{center}
\includegraphics[width=0.5\textwidth]{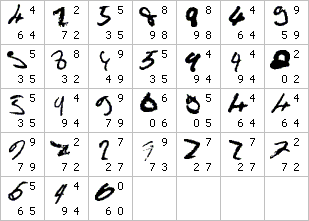}
\end{center}
\caption{The committee's miss-classified digits together with the two most likely predictions (bottom, from left to right) and the correct label (top right) according to MNIST.}
\label{Fig:errors}
\end{figure}

Our GPU implementation of the MLP framework is explained by \cite{Ciresan:2010}. We use the architecture (841 neurons on the input layer, five hidden layers containing 2500, 2000, 1500, 1000 and 500 neurons, and 10 outputs) that has a very low 0.35\% error rate on MNIST. We train six additional nets with the same architecture on preprocessed (normalized) data and form a committee by averaging the predictions of the individual nets. The MNIST data is already preprocessed such that the width or height of the digit is 20 pixels. Variations in writing style result in different aspect ratios of the handwritten digits. We therefore normalize the width of the digits (except for digits 1) to 10, 12, 14, 16, 18 and 20 pixels. Results of the nets trained on normalized data together with the resulting committee are listed in Table~\ref{Table:preprocessed_results}. Interestingly, the error of the committee ($0.31\%$) is considerably lower than those of the individual nets. This is the best result ever reported on MNIST. The 31 misclassified digits are shown in Figure~\ref{Fig:errors}. Many of them are ambiguous and/or uncharacteristic, with obviously missing parts or strange strokes etc. Interestingly, the second guess of the network is correct for 29 out of the 31 misclassified digits.

\begin{table}[h]
	\caption{Error rates of the individual nets and of the resulting committee. WN x---Width Normalization of the bounding box to be x pixels wide; ORIG---original MNIST.}
	\label{Table:preprocessed_results}
	\vspace{8pt}
	\centering
  \begin{tabular}{c|c|c|c}
    net	&	width 	& test error for 				&	best test\\
    		&	normalization			&	best validation [\%]	&	error [\%]\\
    \hline
    1		&	WN 10 						&	0.52									&	0.46	\\
    2		&	WN 12							&	0.45									&	0.37	\\
    3		&	WN 14							&	0.44									&	0.40	\\
    4		&	WN 16							&	0.49									&	0.36	\\
    5		&	WN 18							&	0.36									&	0.31	\\
    6		&	WN 20							&	0.38									&	0.34	\\
    7		&	ORIG							&	0.35									&	0.32	\\
    \hline
    & committee & \bf{0.31}
  \end{tabular}
\end{table}

\section{Conclusion}

Current GPUs are more than 50 times faster than standard microprocessors when it comes to training big and deep neural networks with online back-propagation (weight update rate up to $7.5\times10^9/s$, and more than $10^{15}$ per trained network).  On the competitive MNIST handwriting benchmark, single precision floating-point GPU-based committees of neural nets (each with a a different preprocessor motivated by observed variations in aspect ratio and slant of handwritten digits) outperform all previously published methods, including complex ones involving specialized architectures, unsupervised pre-training, combinations of machine learning classifiers etc.  To avoid overfitting, training sets of sufficient size are obtained by appropriately distorting images.  

% use section* for acknowledgement
\section*{Acknowledgment}

This work was partially funded by the Swiss Commission for Technology and Innovation (CTI),
Project n. 9688.1 IFF: Intelligent Fill in Form.

%\bibliographystyle{plainnat}
%\bibliography{the_bib_jmlr} 

\end{document}